\documentclass[runningheads]{llncs}
\usepackage{graphicx}
\usepackage{hyperref}
\usepackage{amsmath}
\usepackage{amssymb}
\usepackage{cite}
\usepackage{bbold}
\usepackage{rotating}
\usepackage{subcaption}
\usepackage{multirow}
\usepackage{booktabs}
\usepackage{bbm}

\usepackage{academicons}
\begin{document}
\title{Data-Driven Tissue- and Subject-Specific Elastic
Regularization for Medical Image Registration
}
\titlerunning{Tissue- and Subject-Specific Elastic Regularization}
\author{Anna Reithmeir\inst{1,2,3} \and 
Lina Felsner \inst{1,3} \and 
Rickmer Braren \inst{5,6} \and 
Julia A. Schnabel \inst{1,2,3,4} \and  
Veronika A. Zimmer\inst{1,3,5}} 

\authorrunning{A. Reithmeir et al.}


\institute{School of Computation, Information and Technology, Technical University of Munich, Germany \and
Munich Center for Machine Learning, Germany \and
Institute of Machine Learning in Biomedical Imaging, Helmholtz Munich, Germany \and 
School of Biomedical Engineering and Imaging Sciences, King’s College London, UK \and School of Medicine \& Health, Klinkum Rechts der Isar, Technical University of Munich, Germany \and German Cancer Consortium (DKTK), Partner Site Munich, Germany\\*\email{anna.reithmeir@tum.de}}
\maketitle              
\begin{abstract}
Physics-inspired regularization is desired for intra-patient image registration since it can effectively capture the biomechanical characteristics of anatomical structures. 
However, a major challenge lies in the reliance on physical parameters:
Parameter estimations vary widely across the literature, and the physical properties themselves are inherently subject-specific.

In this work, we introduce a novel data-driven method that leverages hypernetworks to learn the tissue-dependent elasticity parameters of an elastic regularizer. Notably, our approach facilitates the estimation of patient-specific parameters without the need to retrain the network.

We evaluate our method on three publicly available 2D and 3D lung CT and cardiac MR datasets. We find that with our proposed subject-specific tissue-dependent regularization, a higher registration quality is achieved across all datasets compared to using a global regularizer. The code is available at \href{https://github.com/compai-lab/2024-miccai-reithmeir}{https://github.com/compai-lab/2024-miccai-reithmeir}.

\end{abstract}

\section{Introduction}
Accurate and reliable intra-patient registration of medical images is ubiquitous for image-based diagnostics and treatment, for example, in radiotherapy and disease monitoring \cite{Rueckert2011}.
While most state-of-the-art deep learning-based registration algorithms resort to generic and global regularization methods (regularizers) \cite{learn2reg2022, VM, lapirn}, 
physics-inspired regularizers, such as the linear elastic regularizer \cite{broit1981}, are particularly well-suited for intra-patient registration. They model biomechanical principles and thus ensure the physical plausibility of the registration.
%
First, the selection of physical parameter values is ambiguous. Prior works often resort to estimations found across the literature \cite{qin2023, minpinn23} even though they can differ significantly not only in their absolute values but also in their ratio \cite{hagemann, talwalkar2008elastography}. 
%
Second, different anatomical structures have different physical properties. For instance, bones are nearly rigid, while soft tissue is more elastic. A global regularizer can not accommodate such varying properties, and thus, a spatially adaptive regularizer that can vary within the image is required. 
However, spatial variability further complicates the choice of physical parameter values, and precise knowledge about anatomical structure locations is necessary.
Third, physical properties can vary among individuals. On the one hand, age is a factor as, e.g., with increasing age, the elasticity of bones and arteries tends to decrease \cite{zioupos2001, cocciolone2018elastin}. On the other hand, the presence of pathologies can impact tissue properties. For example, increased stiffness is observed in fibrotic liver and lung tissue \cite{talwalkar2008elastography, haak2018matrix} as well as in cardiac tissue of patients with obesity and atherosclerosis \cite{cocciolone2018elastin}. 
\begin{figure*}[t]
    \centering
    \includegraphics[width=\textwidth]{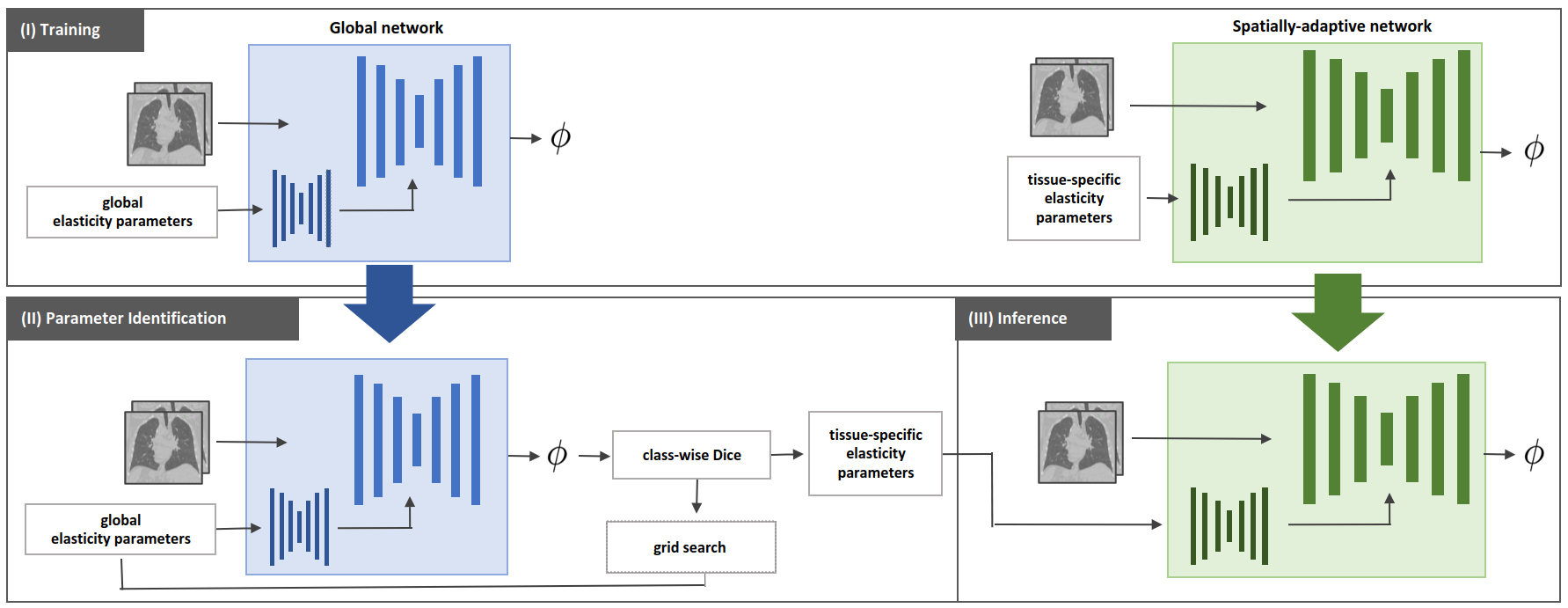}
    \caption{Overview of our method. We train a globally and spatially adaptive network. After training, the optimal tissue-specific elasticity parameters are estimated with the global network. The spatially adaptive network then predicts the deformation field for the registration of a new image pair that follows the physical properties specified by the parameter values.}
    \label{fig:overview}
\end{figure*}
These challenges hinder the use of a one-suits-all set of physical parameter values for the physics-inspired regularization of image registration. 
Recently, works have been introduced that learn the elastic regularization in the context of image registration \cite{ARRATIALOPEZ2023, anonymous}. Arratia L\'opez \textit{et. al.} \cite{ARRATIALOPEZ2023} adopt a hyperelastic regularizer to promote incompressibility in the registration process. While the regularizer varies between the tissue in focus and the background, it does not distinguish between different tissue types. 
Since this method is based on a physics-informed neural network (PINN), retraining is necessary for every image pair.
Reithmeir \textit{et. al.} \cite{anonymous} use a hypernetwork to learn the elasticity parameters of the linear elastic regularizer for image registration. A global regularizer is applied and tissue-specific properties are not considered. 
Neither of the two methods above learns patient-specific regularization parameters. 
The recent work of Ragoza \textit{et. al.} \cite{pinnmiccai23} focuses on the
elasticity parameter estimation from medical images. 
They propose a PINN for spatially varying tissue elasticity reconstruction from images \cite{pinnmiccai23}. However, it requires a specific medical imaging modality (MR elastography) and cannot be applied to standard MR and CT images. Here, image registration is not considered.

In this work, we build upon our previous work in \cite{anonymous} and introduce a novel framework that not only learns tissue- but also subject-specific parameters for the elastic regularization of image registration. 
To this end, we employ two hypernetworks: One for the parameter estimation and one for the registration with the spatially varying elastic regularization (Fig. \ref{fig:overview}). 
%
Our main contributions are twofold:
\begin{itemize}
    \item We propose a data-driven spatially adaptive elastic regularization that varies the elasticity properties for different anatomical structures.
    \item To the best of our knowledge, we are the first to introduce a registration algorithm that \textit{learns} the \textit{subject-specific} linear elastic regularization where the elasticity parameters of the linear elastic regularizer are estimated for each image pair individually without the need to retrain the network.
\end{itemize}

\section{Methods}
\label{sec:method}
We propose a novel method that learns the spatially adaptive linear elastic regularization from the data. It comprises two independent neural networks
that differ in their regularization scheme and which we will refer to as \textit{global} and \textit{spatially adaptive} network in the following (Fig. \ref{fig:overview} (i)).  
The global network employs a global linear elastic regularizer and is used to extract the tissue-dependent elasticity parameters. The spatially adaptive network uses a spatially varying regularizer and is used for the regularized registration. 

For the estimation of optimal tissue- and patient-specific elasticity parameters, we exploit the ability of hypernetworks \cite{ha2017hypernetworks} to adapt to hyperparameter values at test time.
Hypernetworks are small networks that can learn the effect of hyperparameters on the output of a larger network. In this work, we apply two hypernetworks that learn the effect of the elasticity parameters of the linear elastic regularizer \cite{broit1981} on the registration. This allows us to explore the regularization parameter space for new data at test time without the need to retrain the networks.

Our method consists of three stages (see Fig. \ref{fig:overview}), which are described in detail below: (i) Training the two networks, (ii) estimation of the optimal subject- and tissue-specific elasticity parameters, and (iii) registration of new data with the identified parameters. 

\subsection{Training of Global and Spatially Adaptive Networks}
We employ two networks to learn the globally and spatially adaptive regularized registration, respectively. Each consists of a U-Net \cite{ronneberger2015u} registration network and a hypernetwork \cite{ha2017hypernetworks} and predicts the deformation $\phi:\mathbbm{R}^D\rightarrow \mathbbm{R}^D$ between a moving image $M: \Omega \subset \mathbbm{R}^{D}\rightarrow\mathbbm{R}$ and a fixed image $F: \Omega \subset \mathbbm{R}^{D}\rightarrow\mathbbm{R}$. We parameterize $\phi$ as the dense displacement field $\phi(x)=x+\mathbf{u}(x)$ for each pixel location $x$. 

The inputs to both models are the image pair $(M,F)$ and the Lam\'e parameters of the linear elastic regularizer \cite{broit1981} that control the elasticity properties that are modeled. 

\subsubsection{Global Model:} This network is similar to \cite{anonymous} and based on \cite{hypermorph}. It applies a global linear elastic regularizer, i.e., it enforces the same elasticity properties everywhere in the image domain $\Omega$. The two global Lam\'e parameters $\lambda, \mu\in\mathbbm{R}$ are given as inputs to the hypernetwork. 
The network is trained unsupervised, using the loss function in \cite{anonymous}, with the change of constraining $\lambda,\mu\in[0,1]$.
\subsubsection{Spatially adaptive Model:} This network is an extension of \cite{anonymous} with a spatially varying linear elastic regularizer.
It adapts to different elasticity properties based on the underlying tissue type at each $x\in\Omega$. This is achieved by tissue-specific Lam\'e parameters
$\{(\lambda_{i},\mu_{i})|\ \forall i \in 1, ..., C\}$ with $C$ as the number of considered tissue classes. 

Two elasticity parameter maps $\mathbf{\Lambda},\mathbf{\Gamma}:\Omega\in\mathbbm{R}^D\rightarrow\mathbbm{R}$ are formed with the help of the segmentation map $\mathbf{S}:\Omega\subset\mathbbm{R}^D\rightarrow\mathbbm{Z}$ of the fixed image $F$. They hold the respective tissue-specific parameters at each $x\in\Omega$, i.e., $\mathbf{\Lambda}(x)=\lambda_c$ and $\mathbf{\Gamma}(x)=\mu_c$ for $\mathbf{S}(x)=c$.
The training loss function of the network is
\begin{equation} 
\begin{split}
    \mathcal{L}(M,F,\phi, \mathbf{\Lambda}, \mathbf{\Gamma})=& (2*\mathbb{1}-\mathbf{\Lambda}-\mathbf{\Gamma})\ NCC(F,\phi\circ M) \\&+\int_\Omega  \frac{1}{4} \mathbf{\Gamma}\sum_{i,j=1}^{D} (\partial_{x_i}u_j + \partial_{x_j}u_i )^2 + \frac{1}{2} \mathbf{\Lambda} (\mathrm{div\ } \mathbf{u})^2\ d\mathbf{x} \enspace,
\end{split}
\end{equation} 
where $\mathbb{1}:\Omega \subset \mathbbm{R}^{D}\rightarrow\mathbbm{R}$ is a constant weighting map of ones, NCC is the local normalized cross-correlation similarity metric, $\partial_{x}u$ are the partial derivatives of $\mathbf{u}$ and div is the divergence. To balance the regularization and the similarity terms, we require that $0\leq\lambda(x),\mu(x)\leq1\forall x\in\Omega$.

For simplicity, we omit an additional regularization weight which controls the amount of regularization and absorb it in the elasticity parameters as in \cite{anonymous}. During training, the elasticity parameters are individually sampled from the uniform distribution $\mathcal{U}(0,1)$ in each training step. The two networks are trained separately on the same training data. 

\subsection{Elasticity Parameter Identification}
\label{sec:extraction}
Hypernetworks allow the efficient optimization of hyperparameters at test time. We exploit this advantage and use the trained global model to estimate the optimal elasticity parameters of the regularizer. Notably, we do this not only for each tissue type but also for each image pair individually. Thus, we obtain tissue- and subject-specific elasticity parameters, which, to the best of our knowledge, is not performed in prior works.
Similar to \cite{anonymous}, we identify the optimal parameters by a grid search over the parameter space at test time.
%
Without increasing the computational complexity: (i) we perform a sample-based grid search to identify the optimal parameters for each subject individually, and (ii) we base the grid search on a class-wise heuristic to find the optimal parameters for each tissue type. In this work, we use the class-wise Dice score.

Overall, for each test sample $n$, the grid search results in the parameter set $\{(\lambda_{i}^n,\mu_{i}^n)|\ \forall i \in 1, ..., C\}$. For $N$ test samples, in total $2CN$ parameters are estimated. 

\subsection{Registration with Spatially Adaptive Elastic Regularization}
For the registration of an unseen test sample, the optimal Lam\'e parameters that have been identified for this specific sample with the global model are given as input to the hypernetwork of the spatially adaptive model. A single forward pass through the trained spatially adaptive network then registers the given image pair. Since this network is trained with a spatially varying regularizer, the predicted deformation field is regularized in the different anatomical regions according to the specified local elasticity properties. 

\section{Experiments and Results}
\subsection{Implementation Details and Datasets}
We implement our method with TensorFlow 2.12.1 and use the default network architecture of HyperMorph \cite{hypermorph}. We use the Adam optimizer \cite{kingma2014adam} and a learning rate of 1e-4. All networks are trained until convergence using an NVIDIA RTXA6000 GPU. The linear elastic regularizer is implemented with forward finite differences. The code will be made publicly available upon acceptance.

We evaluate our method on three intra-patient datasets in 2D and 3D; (I) \textit{NLST}: The lung CT data of the National Lung Screening Trial dataset provided by the Learn2Reg challenge \cite{NLST, learn2reg2022}. We use the middle coronal slice of each of the 209 annotated images
and a train/val/test split of 169/10/30. We clip the intensities to (-980, 1518) HU.
(II) \textit{ACDC}: The 2D cardiac MRI dataset is taken from the Automatic Cardiac Diagnosis Challenge (ACDC) \cite{acdc}. We use the middle slices of the end-systolic and end-diastolic frames, a split of 50/20/30 and we crop the images to size (128,128).
All images are resampled to isotropic pixel size, and their intensities are normalized.
(III) \textit{L2R-Lung}: The 3D lung CT dataset is taken from the Learn2Reg challenge \cite{lungctdata} and contains 30 inhale/exhale image pairs. The fixed images are only partly visible. 
We use a split of 20/4/4, clip the intensities to (-980, 600) HU and padded the images to size (192,128,192).

In addition to the already provided lung segmentation masks in the NLST and L2R-Lung datasets, we generate liver and bone (vertebrae, ribs, and sternum) masks with the TotalSegmentator tool \cite{totalsegmentator}. The ACDC dataset provides segmentations for the right ventricle, myocardium, and left ventricle blood pool.

\subsection{Analysis of Subject-Specific Elasticity Parameters}
\label{exp1}
\begin{figure}[t]
\centering
\includegraphics[width=.6\textwidth]{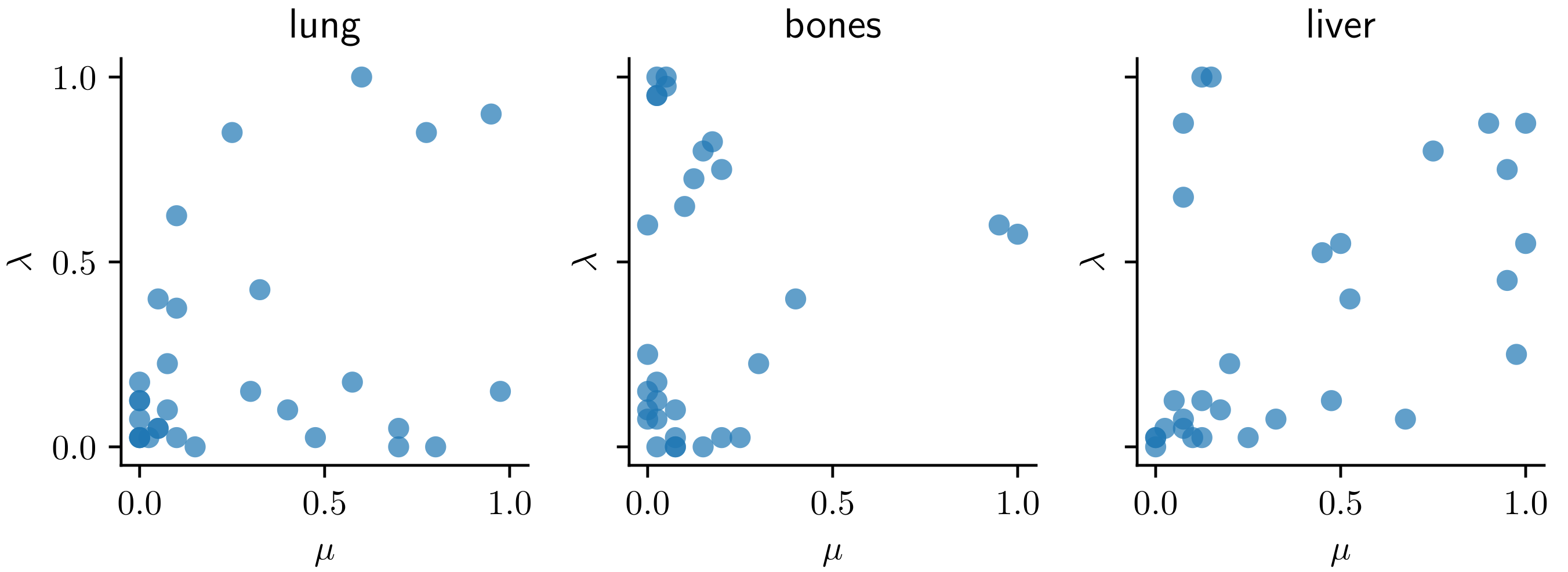}
     \caption{The subject-specific tissue-wise regularization parameters $\lambda,\mu$ that have been estimated with the global network for the NLST dataset. Each blue dot corresponds to one test sample. The results show that the optimal parameters vary between tissue types and subjects.}
     \label{fig:scatter}
\end{figure}

In the first experiment, we analyze the sample-specific elasticity parameters that are estimated with the trained global network as described in Subsection \ref{sec:extraction}. We use the class-wise Dice score as the grid search optimization heuristic and extract the optimal Lamé parameters for each tissue type and subject individually. 
The results for the NLST dataset are shown in Fig. \ref{fig:scatter}. It can be seen that the values for both parameters ($\lambda, \mu$) vary between the subjects. Within each tissue class, a tendency towards low values for $\mu$ and $\lambda$ can be observed. However, for each tissue type, multiple samples can be found for which high values are optimal. Furthermore, the value distribution differs between the tissue types. For example, the bone-specific $\mu$ generally favors values that are smaller than $0.5$, while the liver-specific values stretch out over the full parameter space. Similar results are obtained for the other two datasets (see Appendix A).
\subsection{Comparison of Regularization Methods}
\label{exp2}
In the second experiment, we evaluate our proposed method. To this end, we compare our subject-specific spatially varying regularization (spatially adaptive models) to the global regularization (global models). The global models are evaluated with the estimated optimal global regularization parameters according to the mean Dice score. The spatially adaptive models are evaluated with the estimated optimal tissue-specific regularization parameters according to the class-wise Dice scores. 
Furthermore, for better comparison, we estimate tissue- and subject-specific regularization weights in the same manner for the diffusion regularizer of the HyperMorph model \cite{hypermorph}. Here, the regularization weights specify the amount of smoothness per tissue type. 
As evaluation metrics, we use the mean and class-wise Dice scores (DSC) and the 95th percentile Hausdorff distance (HD95) between the fixed and predicted warped segmentations, the target registration error (TRE, Euclidean distance) between the fixed and predicted warped keypoints and the fraction of negative Jacobian determinants of the displacement field ($\%negJ$). 

\begin{figure}[h!]
\centering
\includegraphics[width=\textwidth]{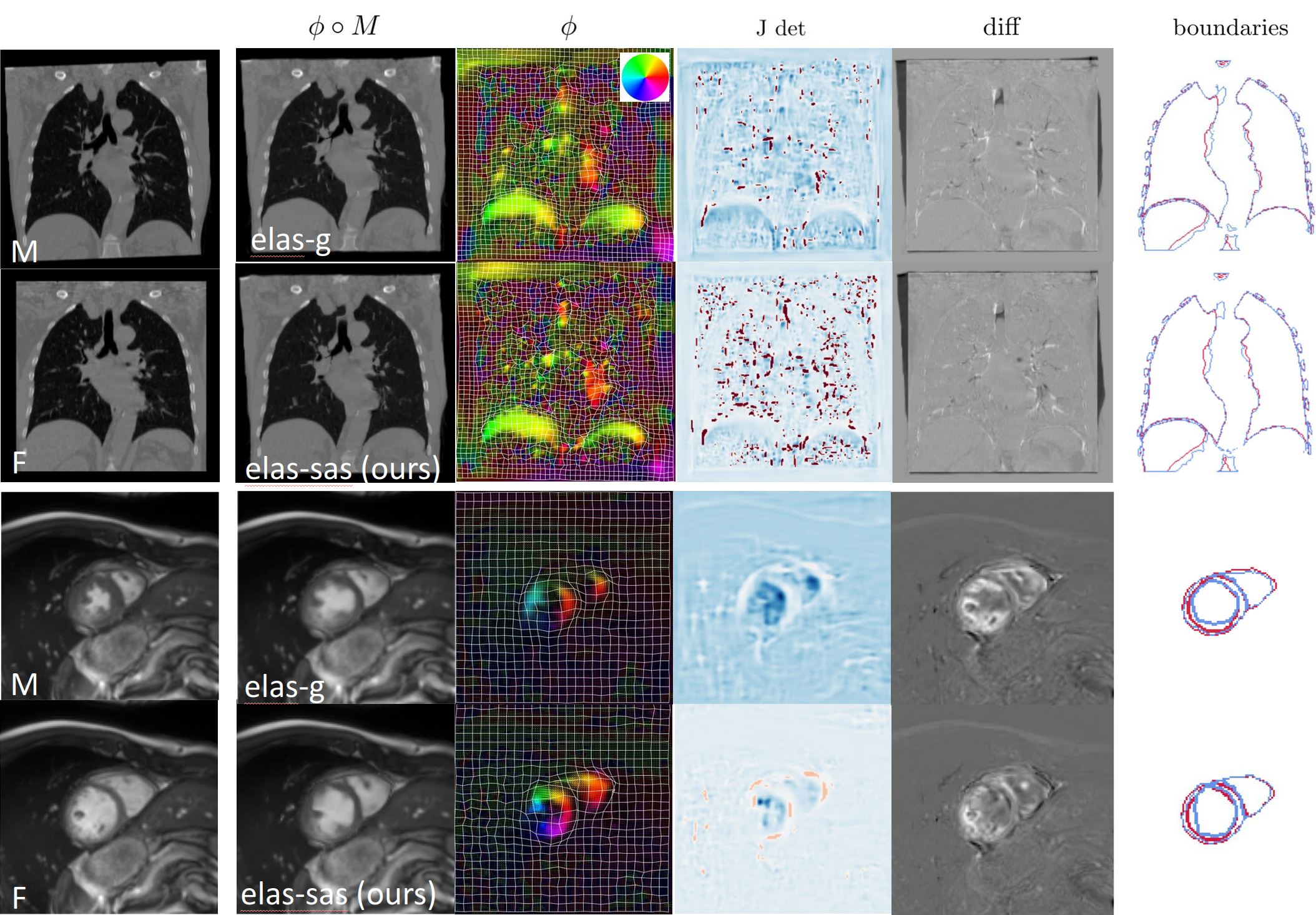}
     \caption{Qualitative results for the NLST (top) and ACDC (bottom) datasets. The left column shows the moving (M) and fixed (F) images. From left to right: Warped moving image, predicted deformation, Jacobian determinant (blue: positive values, red: negative values), difference image between fixed and warped image, segmentation map boundaries. The results show that the subject-specific spatially adaptive regularization (elas-sas) leads to more plausible deformation than the global regularization (elas-g).}
     \label{fig:qualitative}
\end{figure}

The quantitative results are shown in Tab. \ref{tab:results}. 
The results for the diffusion regularizer, in general, outperform the results for the linear elastic regularizer. Similar observations are made in \cite{anonymous}.
In terms of the class-wise DSC, all spatially adaptive models outperform the global models in at least two out of three tissue classes. For the ACDC dataset and the elastic regularizer, all three class-wise DSCs show higher performance with the spatially adaptive regularization, even though this is the only case where a lower HD95 is obtained. 
Furthermore, the mean DSC is consistently higher for the spatially adaptive models. Interestingly, for the 2D datasets, the amount of folding, i.e., the fraction of negative Jacobian determinants, increases when spatially adaptive regularization is applied, while it decreases in the 3D case. Visual inspection of the results reveals that for some test samples, a high amount of folding is observed inside the lung and background (compare second-to-top row of Fig.\ref{fig:qualitative}). Note, however, that nearly no folding is observed here in the liver and bone structures. 
The qualitative results in Fig. \ref{fig:qualitative} furthermore confirm the superior registration quality for the proposed method: The breathing motion of the lower lung lobes (NLST dataset, second-to-top row) and the omnidirectional expansion of the two heart ventricles (ACDC dataset, bottom row) is better reflected in the predicted deformations as can be seen in the deformation plots and difference images. Further qualitative examples can be found in Appendix B.
\begin{table*}[t]
\begin{scriptsize}
    \centering
  \caption{Quantitative results for the linear elastic (\textit{elas}) and diffusion (\textit{diff}) regularized models. The models with global \textit{(g)} and spatially adaptive and subject-specific \textit{(sas)} regularization parameters are compared. The results are shown as mean(standard deviation).  For the class DSC the classes correspond to [lung, bone, liver] for the NLST/L2R-Lung and [right ventricle, myocardium, left ventricle blood pool] for the ACDC dataset. The results show that with the tissue- and subject-specific regularization, higher registration quality is achieved. For metric descriptions see Sec.~\ref{exp2}}
    \begin{tabular}{ccccccc}
    \toprule
        2D &\textbf{Model} & \textbf{Class DSC$\uparrow$} & \textbf{Mean DSC$\uparrow$ }& \textbf{HD95}$\downarrow$ & \textbf{\%negJ$\downarrow$} \\
        \toprule
         \multirow{5}{10pt}{\rotatebox[origin=c]{90}{NLST}}&initial&[0.914, 0.531, 0.775]&0.749(0.064)&11.755(10.328)&-\\
         \cline{2-7}
         &diff-g \cite{hypermorph}&[\textbf{0.972}, 0.799, \textbf{0.855}]&0.885(0.067)&8.227(10.293)&\textbf{0.022}(0.007)\\
         &diff-sas (ours)&[\textbf{0.972}, \textbf{0.801}, \textbf{0.855}]&\textbf{0.886}(0.065)&\textbf{8.157}(9.998)&0.038(0.028)\\
         \cline{2-7}
         &elas-g \cite{anonymous}&[0.969, 0.784, \textbf{0.855}]&0.879(0.068)&9.211(10.636)&\textbf{0.009}(0.004)\\
         &elas-sas (ours)&[\textbf{0.970}, \textbf{0.796}, 0.852]&\textbf{0.883}(0.067)&\textbf{8.598}(10.621)&0.040(0.025)\\
         \bottomrule
         \multirow{5}{10pt}{\rotatebox[origin=c]{90}{ACDC}}&initial&[0.748, 0.520, 0.667]&0.645(0.119)&7.714(3.162)&-\\
         \cline{2-7}
         &diff-g \cite{hypermorph}&[0.897, 0.757, \textbf{0.888}]&0.847(0.052)&2.696(1.776)&\textbf{0.006}(0.005)\\
         &diff-sas (ours)&[\textbf{0.903}, \textbf{0.771}, \textbf{0.888}]&\textbf{0.854}(0.056)&\textbf{2.659}(1.967)&0.011(0.008)\\
         \cline{2-7}
         &elas-g \cite{anonymous}&[0.886, 0.735, 0.867]&0.829(0.066)&\textbf{3.139}(1.682)&\textbf{0.002}(0.002)\\\
         &elas-sas (ours)&[\textbf{0.888}, \textbf{0.743}, \textbf{0.869}]&\textbf{0.833}(0.068)&3.175(2.010)&0.009(0.004)\\
         \toprule
         \toprule
         3D &\textbf{Model} & \textbf{Class DSC$\uparrow$} & \textbf{Mean DSC$\uparrow$ }& \textbf{TRE$\downarrow$/HD95$\downarrow$} & \textbf{\%negJ$\downarrow$} \\
         \toprule
         \multirow{5}{10pt}{\rotatebox[origin=c]{90}{L2R-Lung}}&initial&[0.911, 0.494, 0.758]&0.721(0.061)&3.227(0.734)/16.092(7.328)&-\\
         \cline{2-7}
         &diff-g \cite{hypermorph}&[0.959, 0.693, 0.829] &0.827(0.023)&3.041(0.551)/12.865(7.452)&0.030(0.014)\\
         &diff-sas (ours)&[\textbf{0.960}, \textbf{0.695}, \textbf{0.839}]&\textbf{0.832}(0.022)&\textbf{2.568}(0.532)/\textbf{12.481}(7.579)&\textbf{0.027}(0.010)\\
         \cline{2-7}
         &elas-g \cite{anonymous}&[\textbf{0.961}, 0.678, 0.822]&0.820(0.022)&2.645(0.484)/13.330(7.942)&0.013(0.005)&\\
         &elas-sas (ours)&[0.959,\textbf{0.684},\textbf{0.842}]&\textbf{0.828}(0.024)&\textbf{2.638}(0.512)/\textbf{12.499}(6.786)&\textbf{0.006}(0.002)\\
         \bottomrule
    \end{tabular}
    \label{tab:results}
    \end{scriptsize}
\end{table*}

\section{Discussion and Conclusion}
In this work, we proposed a novel method that estimates tissue- and subject-specific parameters for the physics-inspired regularization of data-driven image registration. We focused on the parameters of the linear elastic regularizer and evaluated the method on three diverse datasets. 

Our analysis of the subject-specific elasticity parameters that have been estimated with the global model in Sec.\ref{exp1} confirms that the parameters can differ not only between tissue types but also between individual subjects. This highlights the importance of the adaptation of physics-inspired regularizers to the local anatomical structures and to the individual patients. The comparison of the different regularization methods in Sec.\ref{exp2} shows that physiologically more meaningful deformations are obtained with spatially adaptive, subject-specific linear elastic regularization. While our method results in more folding, we observe consistently increased class Dice scores for the bones in Tab.\ref{tab:results}, which indicates that small anatomical structures are better registered with our method than with a globally regularized model. 

A current limitation of this work is that the estimated elasticity parameters cannot be interpreted in a physically meaningful manner. The parameter $\mu$ is the shear modulus and serves as a measure of stiffness: Rigid materials have higher values for $\mu$ than more elastic materials. This, however, is not reflected in our estimations since, e.g., the bone-specific $\mu$ do not favor larger values compared to the liver- or lung-specific $\mu$. A possible reason for this could be the uniform sampling space of the parameters in our method, and we plan to investigate this further in the future. Also, we aim to extend the evaluation of our method to additional datasets and more tissue types. 

In conclusion, we proposed a novel image registration method that learns tissue- and sample-specific elasticity parameters for physics-inspired regularization. We believe that the patient-specific adaptation of the regularization is fundamental for successful intra-patient registration and that it can further improve data-driven registration methods in the future.


\clearpage
%
%
%
\bibliographystyle{splncs04}
\bibliography{bib}

\newpage
\section*{Supplementary Material}
\subsection*{Appendix A}
\begin{figure}[h]
\center
\begin{subfigure}{.75\textwidth}
    \includegraphics[width=\textwidth]{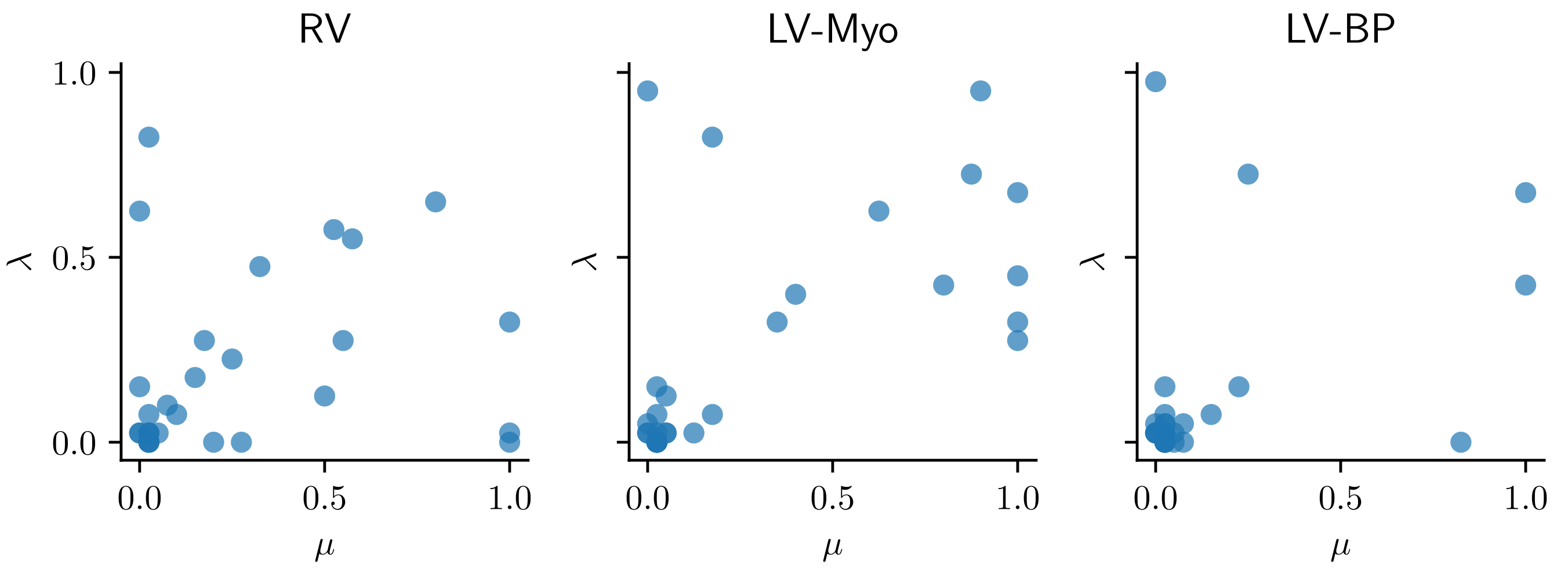}
    \caption{ACDC dataset}
    \label{fig:first}
\end{subfigure}
\begin{subfigure}{.75\textwidth}
    \includegraphics[width=\textwidth]{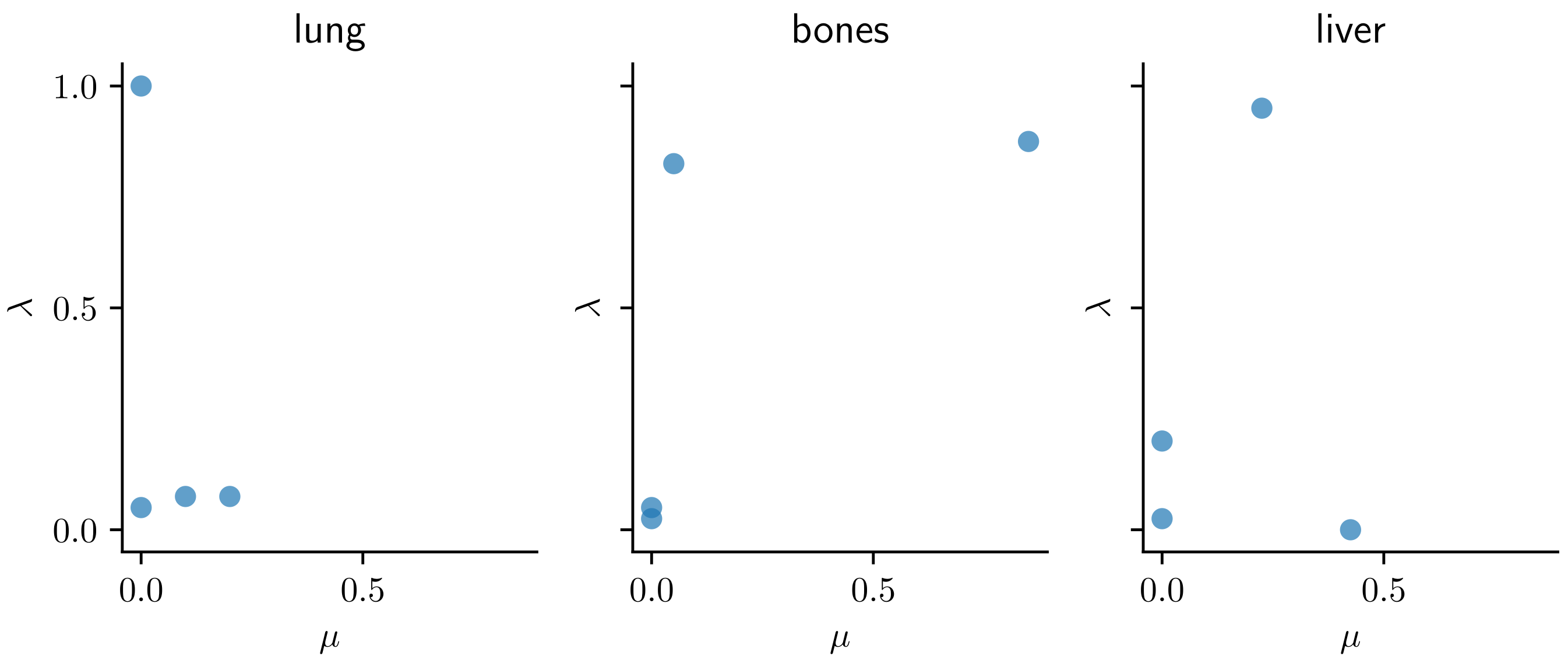}
    \caption{L2R-Lung dataset}
    \label{fig:second}
\end{subfigure}
\caption{The subject-specific tissue-wise regularization parameters $\lambda,\mu$ for the ACDC and L2R-Lung datasets that are estimated with the trained global model. In Fig.~\ref{fig:first}, the classes correspond to the right ventricle (RV), myocardium (LV-Myo), and left ventricle blood pool (LV-BP).}
\end{figure}

\newpage
\subsection*{Appendix B}
\begin{figure}[h]
\centering
\includegraphics[width=\textwidth]{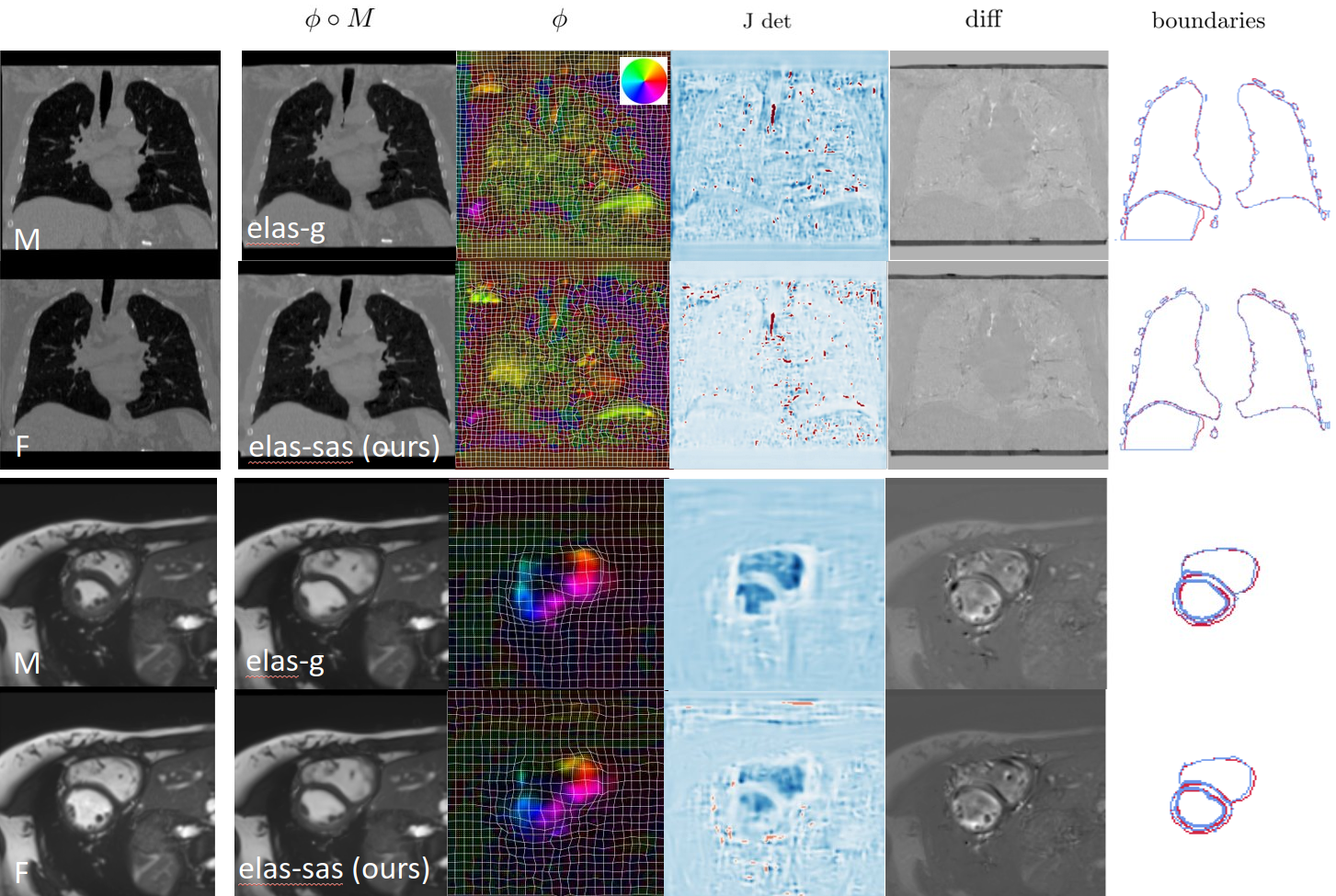}
     \caption{Further visual registration examples of the NLST (top) and ACDC (bottom) test dataset. The left column shows the moving (M) and fixed (F) images. From left to right: Warped moving image, predicted deformation, Jacobian determinant (blue: positive values, red: negative values), difference image between fixed and warped image, segmentation map boundaries.}
     \label{fig:second}
\end{figure}

\end{document}